\documentclass[10pt,twocolumn,letterpaper]{article}

\usepackage{wacv}
\usepackage{times}
\usepackage{epsfig}
\usepackage{graphicx}
\usepackage{amsmath}
\usepackage{amssymb}
\usepackage{hhline}
\usepackage{array,multirow}

\DeclareMathOperator*{\argmin}{arg\,min}
\def\a{\boldsymbol{a}}
\def\g{\boldsymbol{g}}

\usepackage{bigstrut}
\newlength\mylena
\newlength\mylenb
\newcommand\mystrut[1][2]{%
    \setlength\mylena{#1\ht\@arstrutbox}%
    \setlength\mylenb{#1\dp\@arstrutbox}%
    \rule[\mylenb]{0pt}{\mylena}}

\wacvfinalcopy 

\def\wacvPaperID{870} 

\ifwacvfinal\fi
\begin{document}

\title{Boosting Deep Face Recognition via Disentangling Appearance and Geometry}

\author{Ali Dabouei, Fariborz Taherkhani, Sobhan Soleymani Jeremy Dawson, Nasser M. Nasrabadi\\
West Virginia University\\
{\tt\small ad0046@mix.wvu.edu, fariborztaherkhani@gmail.com, ssoleyma@mix.wvu.edu}\\
{\tt\small\{nasser.nasrabadi, jeremy.dawson\}@mail.wvu.edu }}

\maketitle
\ifwacvfinal\thispagestyle{empty}\fi

\begin{abstract}
In this paper, we propose a framework for disentangling the appearance and geometry representations in the face recognition task. To provide supervision for this aim, we generate geometrically identical faces by incorporating spatial transformations. We demonstrate that the proposed approach enhances the performance of deep face recognition models by assisting the training process in two ways. First, it enforces the early and intermediate convolutional layers to learn more representative features that satisfy the properties of disentangled embeddings. Second, it augments the training set by altering faces geometrically. Through extensive experiments, we demonstrate that integrating the proposed approach into state-of-the-art face recognition methods effectively improves their performance on challenging datasets, such as LFW, YTF, and MegaFace. Both theoretical and practical aspects of the method are analyzed rigorously by concerning ablation studies and knowledge transfer tasks. Furthermore, we show that the knowledge leaned by the proposed method can favor other face-related tasks, such as attribute prediction.    
\end{abstract}

\section{Introduction}
Using the face as a biometric trait has several advantages for identification purposes. First, faces are naturally exposed and can be often captured remotely with suitable quality by incorporating a ubiquitous, moderately priced camera system. Second, the convenience of the acquisition procedure has promoted the acceptability of the modality compared to the fingerprint and iris which require direct cooperation of individuals. Third, the consistent morphological structure of faces, \ie, semantic parts of the face share similar spatial properties among different individuals, also facilitates the process of reducing the variations of face images captured in unconstrained setups. In classical face recognition (FR) studies, the major challenge was to devise hand-crafted features that offer high inter-class separability and low intra-class variations \cite{ahonen2006face, belhumeur1997eigenfaces, duan2018context, lei2013learning}.

\begin{figure}[t]
    \centering
    \includegraphics[width=0.4\textwidth]{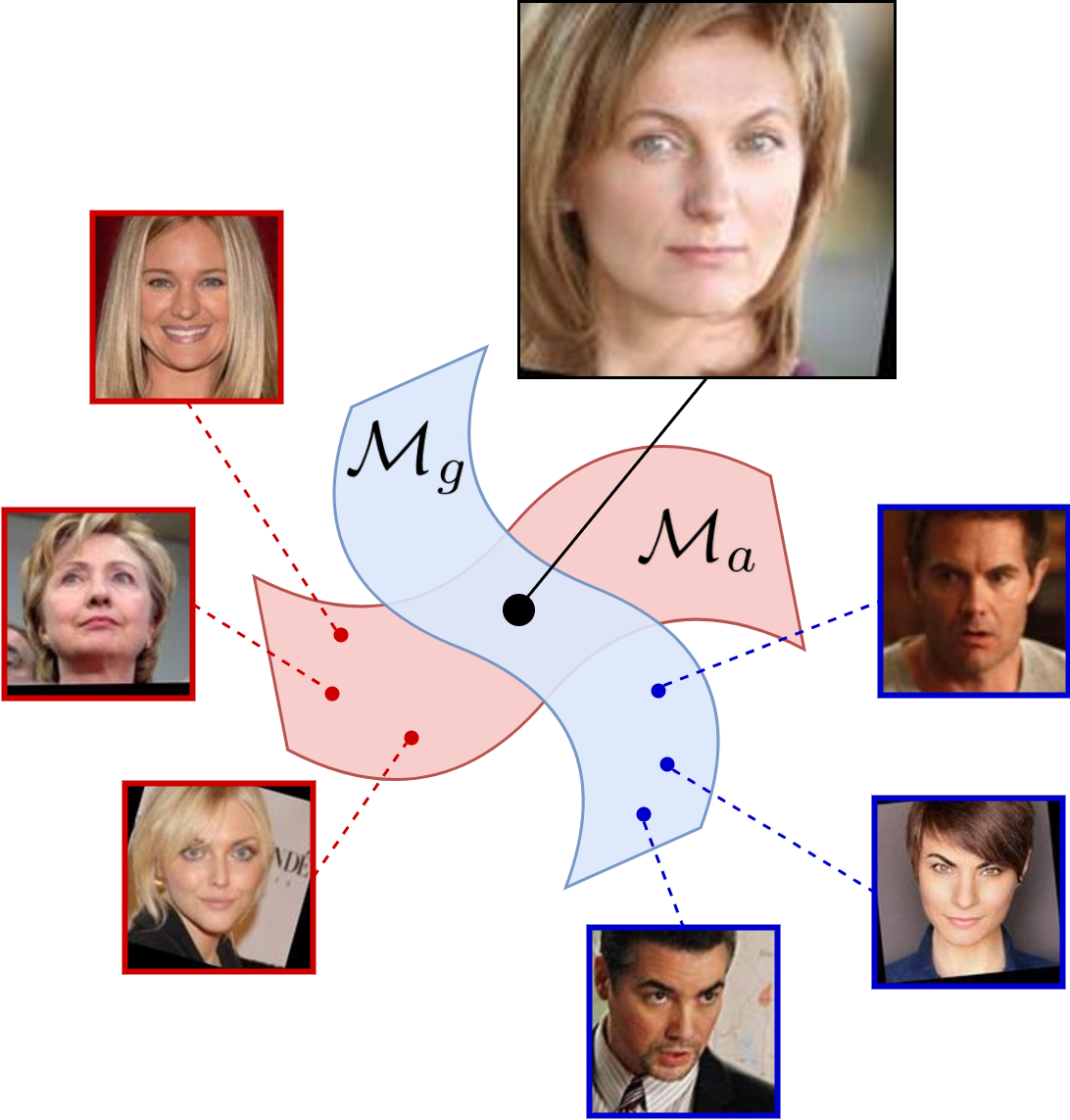}
    \caption{The proposed approach disentangles deep representations of the appearance and geometry of the face. 
    $\mathcal{M}_g$ and $\mathcal{M}_a$ provide a schematic visualization of the manifolds of appearance and geometry constructed using our framework, respectively. Manifolds are superimposed at the input face representation.}
    \label{fig:my_label}
\end{figure}

The rapid development of technology over the last decade has had a profound impact on the performance and methodology of FR approaches. It has led to the generation of large-scale face datasets such as VGGFace \cite{parkhi2015deep}, CASIA-WebFace \cite{yi2014learning}, and MS-Celeb-1M \cite{guo2016ms}. Such comprehensive datasets of faces revealed detailed information regarding the manifold of natural faces and provided the supervision for training large-scale learning models. On the other hand, the development of parallel processing units has allowed an efficient optimization of deep neural networks (DNNs) which consist of millions of trainable parameters. Consequently, the classical problem of FR has transformed into the new challenge of finding efficient and powerful network architectures and suitable loss functions. To this aim, a myriad of approaches has been proposed to learn discriminative face representations using DNNs \cite{parkhi2015deep, schroff2015facenet, sun2014deep, wang2018cosface, liu2017sphereface}.

Most recently, spectacular performance of carefully designed network architectures, such as VGG \cite{parkhi2015deep} and ResNet \cite{he2016deep}, have concentrated attention on finding suitable loss functions and training criteria \cite{taherkhani2019matrix, taherkhani2019weakly}. A suitable criterion should force the model to learn discriminative representations for which the maximum intra-class distance is smaller than the minimum inter-class disparity \cite{liu2017sphereface}. However, the challenging effects of unconstrained environmental conditions, such as lightning and backgrounds, in addition to the intrinsic variations of human pose and facial expressions, complicates finding a suitable criterion. Contrary to the object recognition problem, in the FR task, the number of classes is extremely large and the number of available samples per each class is often limited and variable. This significantly degrades the performance of the well-established Softmax loss function, \ie, the combination of the Softmax normalization and cross-entropy loss function. Indeed, Softmax loss yields separable features but cannot provide sufficient discrimination \cite{liu2017sphereface}. Pioneer deep learning-based FR approaches have sought to increase the discrimination power of deep representations by devising novel losses, such as contrastive loss \cite{sun2014deep}, center loss \cite{wen2016discriminative} and triplet loss \cite{schroff2015facenet}. However, recent studies have demonstrated that considering angular distance instead of the euclidean distance for the Softmax loss significantly improves the discrimination power of representations \cite{liu2017sphereface, wang2018cosface}. Hence, Softmax loss refined by angular distance has become the state-of-the-art method for training deep models.   

In this paper, we seek to improve the performance of deep FR models by considering a novel perspective: instead of modifying the classification loss functions to obtain compact and discriminative representations, we propose disentangling of the appearance and geometry of the face. The core idea of the paper is to construct geometrically identical faces by incorporating spatial transformations and exploiting their relative similarities to learn disentangled embedding representations. Practically, the disentanglement provides two benefits for the training procedure of deep FRs. First, it improves the generalization and training accuracy by geometrically augmenting the training set. Second, it enhances the learned knowledge of the early and intermediate layers of the deep model by enforcing them to satisfy the relative properties of appearance and geometry representations in the corresponding embedding spaces. We conduct extensive experiments to evaluate these benefits for the face recognition task and demonstrate that the knowledge learned through the disentangling approach can also be used to improve the performance of other face-related tasks, such as attribute prediction. 

\section{Related Work}

\noindent{\bf Classical face recognition.} Face recognition has always been an important computer vision problem due to its challenging aspects, such as the large number of classes and limited number of per-class samples. Classical approaches have mainly addressed the problem by finding discriminative hand-crafted representations for the face. Most of the attempts were strongly dependent on experimental observations since the prior knowledge needed for extracting hand-crafted features were scarce or hard to interpret. Besides, capturing large intra-class variations was also a major hurdle. Similar to the hierarchy of cascaded computations in DNNs, hand-crafted methods, such as LBP \cite{ahonen2006face} and Gabor \cite{liu2002gabor}, compute local descriptors and combine them to obtain higher-level representations with more discriminative power. However, these heuristic methods can offer limited discrimination since they are not directly supervised to optimize the classification objective \cite{cao2010face}. In addition, although they are data-independent, they cannot robustly capture intra-class variations.    

\vspace{2pt}
\noindent{\bf Deep face recognition.} In recent years, DNNs have achieved astonishing performance in face recognition which has gone even beyond human-level expertise. As the pioneers of the work, DeepFace \cite{taigman2014deepface} and DeepID \cite{sun2014deep} incorporated the well-known combination of Softmax normalization and cross-entropy loss for learning very deep representations of the face. These were accompanied by studies expanding the network architectures and gathering large-scale datasets, such as VGG \cite{parkhi2015deep}. Although Softmax loss provides separability of classes, the learned features are not discriminative enough. Hence, several novel training criteria and loss functions have been proposed to enhance the discrimination power of learned representations. Sun \etal \cite{sun2014deep} incorporated a verification loss to enhance the performance of the identification loss. Schroff \etal \cite{schroff2015facenet} proposed a novel training criterion called triplet loss in which the representations are forced to be discriminative based on the relations of a triplet of embedding samples. Wen \etal \cite{wen2016discriminative} proposed center loss to increase the intra-class compactness of representations. Finally, based on the observation that Softmax loss imposes an angular distribution on the representations, several studies have proposed to enhance the discrimination power of representations by mapping faces onto hyperspherical embeddings and measure their similarities using the Cosine measure \cite{liu2017sphereface, wang2018cosface}.  

\vspace{2pt}
\noindent{\bf Disentangling geometry and appearance of the face.} Geometry and appearance are the two main characteristics of the face which are highly correlated with the corresponding ID. Since the very first research on FR, the geometry is known to play a crucial role in identification \cite{galton1889personal, iranmanesh2020robust,dabouei2019fast}. This has also been exploited to find suitable hand-crafted features for face recognition \cite{ahonen2006face}. Appearance is a major part of a general term called soft biometrics which encompasses all characteristics of individuals which do not need to be unique but can help recognize the ID, \eg, hair color and gender. Several approaches have been considered in soft biometrics to enhance the FR performance \cite{jain2009facial, park2010face}. An important limitation of these studies is their dependence on the soft biometrics information in the dataset. They also require appearance information during the test phase.         

Several prior studies attempted to disentangle the appearance and geometry of faces. Shu \etal \cite{shu2018deforming} proposed an unsupervised approach by using a coupled autoencoder model. Each of the autoencoders is forced to learn the geometry or appearance representation of the input sample. The model provides the supervision for disentangling by reconstructing the original image using the combination of the two representations. Xing \etal \cite{xing2018deformable} followed a similar methodology but incorporated variational autoencoders to enhance the performance of disentangling. These methods provided a novel insight toward the task. However, representations learned using autoencoders do not contain enough identification information to achieve state-of-the-art performance in face recognition. 

\section{Disentangling Geometry and Appearance}
In this study, the main supervision for disentangling appearance and geometry of faces is provided by constructing two pairs of face images. In the first pair, the appearance of faces is similar and the geometry is different, while, in the second pair, the geometry is similar and the appearance is different. For this purpose, we geometrically map an input face image to another ID in the training set using landmark information available for face alignment. The combination of the manipulated face image with its original version and the target face image construct the first and second pairs of faces, respectively.   
\begin{figure}
    \centering
    \includegraphics[width=0.46\textwidth]{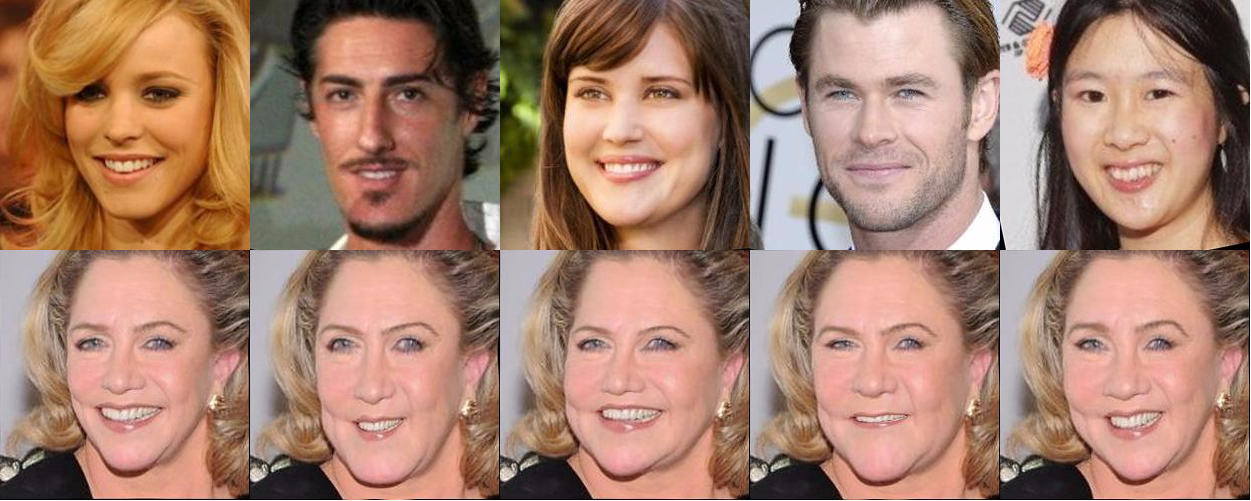}
    \caption{Examples of geometrically identical faces generated for five different IDs. First row shows input faces $x_i$, and the second row shows the corresponding face images $\hat{x}_{i'}$. }
    \label{fig:geometrymapsamples}
\end{figure}
\subsection{Geometrically Identical Faces}\label{sec:geoidenticalfaces}
Let $x_i$ be an input face image belonging to class $y_i$ and the set $l_i = \{(u_j, v_j): \forall j\in \{1,\dots, K\}\}$ describe the 2D locations of $K$ landmarks corresponding to $x_i$. For each input face image, we find the closest neighbor face, $x_{i'}$, from a different class, $y_{i'}\neq y_i$, in the geometry space by computing $i'$ as: 
\begin{equation}\label{eq:neighbor}
    i' = \argmin\limits_j \dfrac{||l_i - l_j||_\infty}{\delta(y_i - y_j)+\epsilon}  ,
\end{equation}
where $\delta(.)$ is one when the input argument is not  zero and is zero otherwise, and $\epsilon \ll 1$, \eg $10^{-6}$. Here, $\ell_\infty$-norm assures that the selected neighbor face has a similar structure and pose as the input image in order to minimize the distortion caused by the spatial transformation in the next step. We assume that the rotation, scale, and translation of faces are aligned for the whole dataset, thus, the similarity of $l_i$ and $l_j$ can be measured in the same frame. It also worth mentioning that, for this work, we assume all landmark locations are vectorized before performing $\ell_p$-norm, $||\cdot||_p$.  

Although face image $x_{i'}$ has a geometry similar to $x_i$, their geometries do not completely match. To further match the geometry of two samples, we incorporate a spatial transformation and map $x_{i'}$ to $x_{i}$ such that the resulting image has the same set of landmark locations. The deformed face image can be computed as:
\begin{equation}\label{eq:spatialtransformation}
    \hat{x}_{i'} = T(x_{i'}, l_{i'}, l_i),
\end{equation}
where T is the spatial transformation, \ie thin plate spline (TPS) \cite{bookstein1989principal}, which has a suitable capacity for the desired mapping compared to the affine transformation. The resulting face image, $\hat{x}_{i'}$, has the geometry of $x_i$ and the appearance of $x_{i'}$. Figure \ref{fig:geometrymapsamples} shows examples of this mapping. It may be noted that one can geometrically map all faces in the dataset to a canonical template in order to provide the supervision for decomposing the appearance and geometry of faces. However, computing a geometrically identical face for each input face provides two major benefits. First, it augments the training set by geometrically manipulating face images. Second, it increases the performance of the spatial transformer in matching the geometry of faces since each face is mapped to a face which is geometrically similar. In the next subsection, we use the geometric similarity and appearance disparity of $x_i$ and $\hat{x}_{i'}$ as the main supervision for the disentangling process.   

\subsection{Disentangling Networks}
We define two networks for learning the discriminative representations of geometry and appearance of faces. Let $\g: \mathbb{R}^{w\times h \times 3} \rightarrow \mathbb{R}^{d_g}$ be the function mapping input image $x$ to the geometric representation of the input face with the cardinality $d_g$. Also, let $\a: \mathbb{R}^{w\times h \times 3} \rightarrow \mathbb{R}^{d_a}$ be the function mapping the same face to the representation of the appearance. For brevity, we assume the cardinality of both representations are equal $d_g = d_a = d$. We also define a third function $f: \mathbb{R}^{d}\times \mathbb{R}^{d} \rightarrow \mathbb{R}^{d'}$ which takes the geometry and appearance representations and maps them to the final $d'$-dimensional representation of the input face.

Based on the properties of geometrically identical faces defined in Subsection \ref{sec:geoidenticalfaces}, representations of the appearance and geometry of the faces $x_i$, $x_{i'}$, and $\hat{x}_{i'}$ should satisfy following conditions: i) the geometry representations of the input face and the manipulated face should be equal, $\g(x_i) \approx \g(\hat{x}_{i'})$, ii) the appearance representations of the neighbor face and its transformed version should be equal, $\a(x_{i'}) \approx \a(\hat{x}_{i'})$, and iii) integrating representations using $f$ should provide enough information for an accurate identification of the input samples $x_i$ and $x_{i'}$. We define proper loss functions to enforce such conditions on the representations. Intrinsically, the conventional Softmax loss function imposes an angular distribution on the learned representations \cite{wen2016discriminative}. Hence, we use the cosine similarity, which is a more suitable metric compared to Euclidean distance, to define the loss functions. As a result, representations of the appearance, geometry, and final ID in our framework follow an angular distribution.  
\begin{figure}[t]
    \centering
    \includegraphics[width=0.46\textwidth]{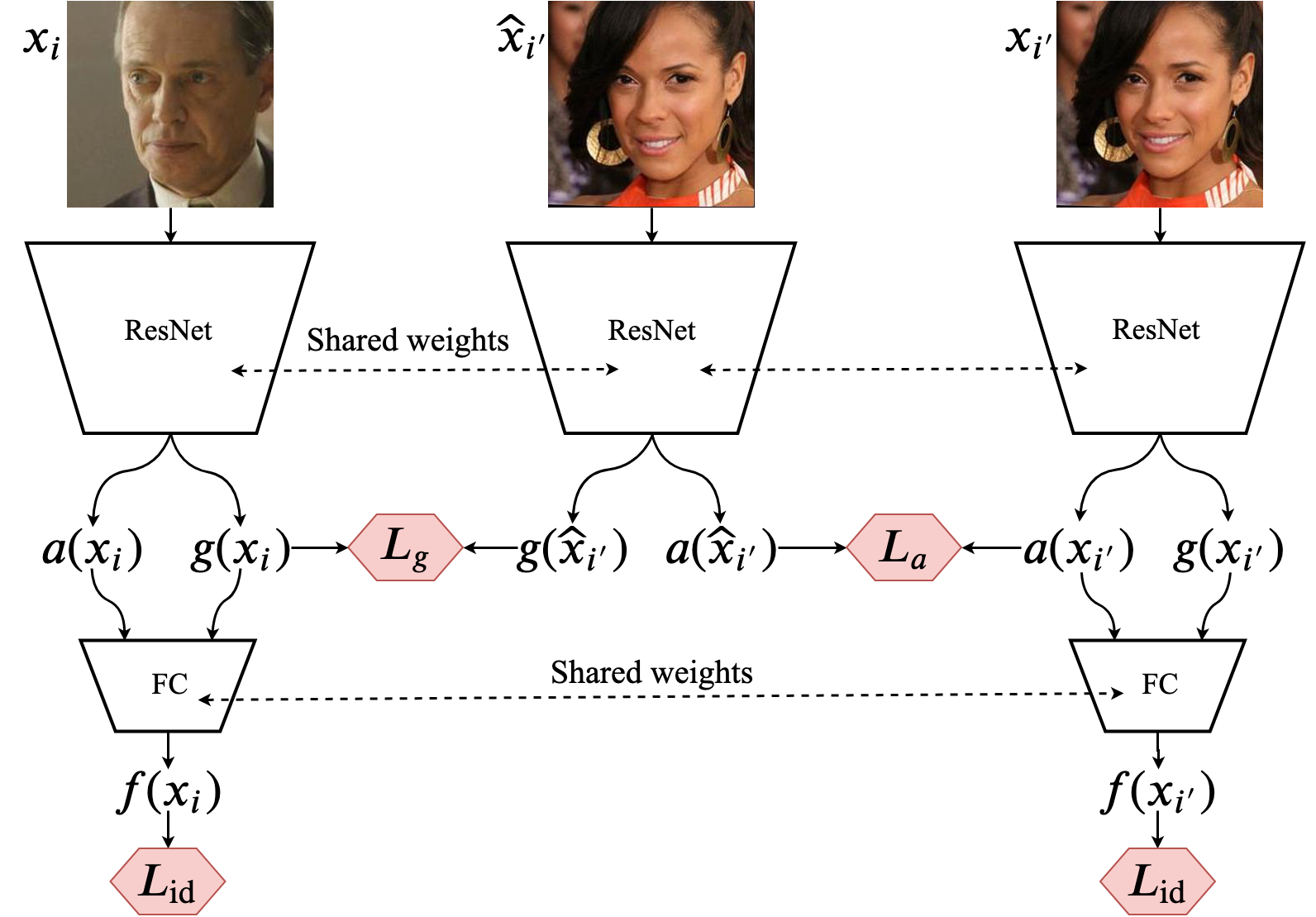}
    \caption{Schematic for training face recognition models enhanced by the proposed DAG approach.}
    \label{fig:disentanglingdiag}
\end{figure}

\begin{figure*}[t]
    \centering
    \includegraphics[width=0.98\textwidth]{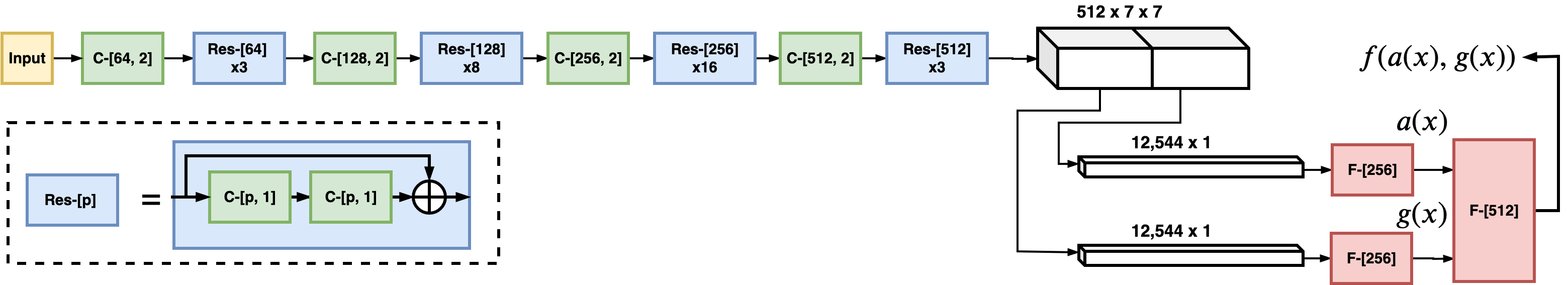}
    \caption{Architecture of the disentangling model. The network comprised of 3x3 convolutions consisting of p filters with the stride of s followed by PReLU (C-[p, s]), and fully-connected with p outputs (F-[p]). Each ResNet block consists of two consecutive convolutional layers followed by a shortcut from its input to its output.  }
    \label{fig:netarch}
\end{figure*}

We satisfy the first condition by defining a geometry-preserving loss function as: 
\begin{equation} \label{eq:contrastivegeometry}
    \begin{split}
          \mathcal{L}_g := -\dfrac{1}{N}\sum\limits_{i} \Phi(\g(x_i), \g(\hat{x}_{i'}))~~~~~~~~~~~~~\\+\max(0, \Phi(\g(x_i), \g(x_{i'}))-\alpha_g\phi_g), 
    \end{split}
\end{equation}
where $\Phi(v_1, v_2) = \tfrac{v_1^Tv_2}{||v_1||||v_2||}$ computes the cosine similarity of input vectors, and $N$ is the number of total samples in the batch. $\phi= \tfrac{||l_i - l_{i'}||_2}{||l_i-\overline{l}_i||_2}$ is a normalized measure of the distance of landmark locations, and $\overline{l}_i$ is the mean of landmark locations along two axes. $\alpha_g$ is a coefficient scaling $\phi_g$ to construct a margin which controls the angular distance between the geometry representation of $x_i$ and $x_{i'}$.   
Indeed, Equation \ref{eq:contrastivegeometry} forms an angular contrastive loss which aims to maximize the cosine similarity of $\g(x_i)$ and $\g(\hat{x_{i'}})$ while assuring that $\g(x_i)$ and $\g(x_{i'})$ are dissimilar, proportional to the landmark disparity of $x_i$ and $x_{i'}$.   
    
Similarly, we define the appearance-preserving loss function as: 
\begin{equation}\label{eq:apearance}
          \mathcal{L}_a := -\dfrac{1}{N}\sum\limits_{i} \Phi(\a(x_{i'}), \a(\hat{x}_{i'})).
\end{equation}
Face samples $x_i$ and $x_{i'}$ are selected solely based on their geometric similarity, and their appearance can be completely different or very similar. Hence, Equation \ref{eq:apearance} does not consider a contrastive loss as the similarity of $\a(x_i)$ and $\a(x_{i'})$ is ambiguous. However, the identification loss function, described in the following, encourages appearance representations of different faces to take large enough distances for providing accurate identification. 

So far, we developed a technique for {\bf D}isentangling the {\bf A}ppearance and {\bf G}eometry ({\bf DAG}) representations of face images. The final step is to combine this approach with conventional FR methods to establish highly discriminative representations. To this aim, we combine DAG with the family of A-Softmax \cite{liu2017sphereface, wang2018cosface} loss functions which have demonstrated significant performance for face recognition task. The main formulation of the loss function is: 
\begin{equation}\label{eq:angualrloss}
    \begin{split}
    \mathcal{L}_{id} = \dfrac{-1}{N}\sum\limits_{i} \log  \dfrac{e^{s(\cos(m_1\theta_{{y_i}, i})-m_2)}}{e^{s(\cos(m_1\theta_{{y_i}, i})-m_2)}\!+\!\sum\limits_{j\neq y_i} e^{s\cos(\theta_{{j}, i})}},
    \end{split}
\end{equation}
where $z_i = f(\a(x_i), \g(x_i))$ is the final representation obtained by combining geometry and appearance information. Here, $\cos(\theta_{j, i})=\tfrac{1}{||z_i||||W_i||}W_i^Tz_i$, where $W_i$ is the weight vector assigned to the $i^{th}$ class. Variables $m_1$ and $m_2$ are hyper-parameters controlling the angular margin, and $s$ is the magnitude of angular representations. $\mathcal{L}_{id}$ with $(m_1=4, m_2=0, s=||z_i||)$ and $(m_1=0, m_2=0.35, s=64)$ denotes the loss functions defined in SphereFace \cite{liu2017sphereface} and CosFace \cite{wang2018cosface}, respectively. Later in Section \ref{sec:facerecongitionenhancedbydag}, we combine both these loss functions with DAG to evaluate the effectiveness of the integrated model. The total loss for training the proposed framework is $\mathcal{L}_t = \mathcal{L}_{\overline{id}} + \lambda_a\mathcal{L}_a + \lambda_g\mathcal{L}_g$, where $\lambda_a$ and $\lambda_g$ are hyper-parameters scaling the appearance and geometry preserving loss functions. Furthermore, $\mathcal{L}_{\overline{id}}$ is the average recognition loss function on $x_i$ and $x_{i'}$. Figure \ref{fig:disentanglingdiag} presents a schematic of the training criteria.    

\section{Experiments}
Here, we evaluate the effectiveness of the proposed disentangling approach. First, we describe the implementation setup of the proposed model in Subsection \ref{sec:implementationdetail}. Afterward, we conduct exploratory experiments to tune the hyper-parameters and provide some visualizations of the learned embedding representations in Subsection \ref{sec:lambda}. Finally, we evaluate the performance of the face recognition and attribute prediction tasks enhance by DAG in Subsections \ref{sec:facerecongitionenhancedbydag} and \ref{sec:dagforattributeprediction}, respectively.   
\subsection{Implementation Details}\label{sec:implementationdetail}
\noindent{\bf Architecture and Hyper-parameters:} We adopt ResNet \cite{he2016deep} for the base network architecture of our model. To reduce the size of the model, the convolutional networks for extracting the geometry representation, $\g(x)$, and the appearance representation, $\a(x)$, are combined in a single ResNet-64 architecture. This network produces feature maps of spatial size $7\times 7$ and the depth of $512$ channels. Feature maps are then divided in depth into two chunks, and the first and second chunks are dedicated to the appearance and geometry, respectively. Feature maps are then reshaped to form vectors of size $12,544$ and passed to dedicated fully-connected layers to generate the final representations of the appearance, $\a(x)$, and geometry $\g(x)$. The cardinality of geometry and appearance representations is set to $d=256$. The linear mapping $f$ takes the concatenated outputs of $\g$ and $\a$ and maps them using a fully connected layer to the final embedding with the cardinality $d' = 512$. Figure \ref{fig:netarch} details the network architecture of the model.

\begin{figure}
    \centering
    \includegraphics[width=0.35\textwidth]{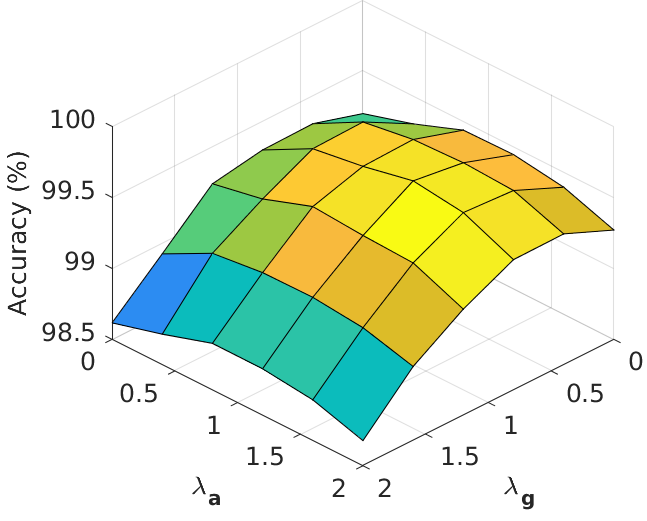}
    \caption{Accuracy (\%) of Softmax loss enhanced by DAG trained with different values of $\lambda_a$ and $\lambda_g$ on LFW \cite{huang2008labeled}.}
    \label{fig:lambdavslambda}
\end{figure}
The model is trained using Stochastic Gradient Descent (SGD) with the mini-batch size of $128$ on four NVIDIA TITAN X GPUs. The initial value for the learning rate is set to $0.1$ and multiplied by $0.9$ in intervals of five epochs until its value is less than or equal to $10^{-6}$. All models are trained for 600K iterations. The average of the landmark disparity measure $\phi_g$ on the training set of CASIA-WebFace \cite{yi2014learning} is $\approx 0.103$. Accordingly to this value and based on practical evaluations, we found that $\alpha_g=9.4$, \ie keeping the angle of $\g(x_i)$ and $\g(x_{i'})$ greater than $\tfrac{\pi}{9}$, yields significant discriminability of geometry representations. We also set $\lambda_a=1.3$ and $\lambda_g=0.75$ based on experiments conducted in Section \ref{sec:lambda}.
\begin{figure*}[t]
    \centering
    \includegraphics[width=0.99\textwidth]{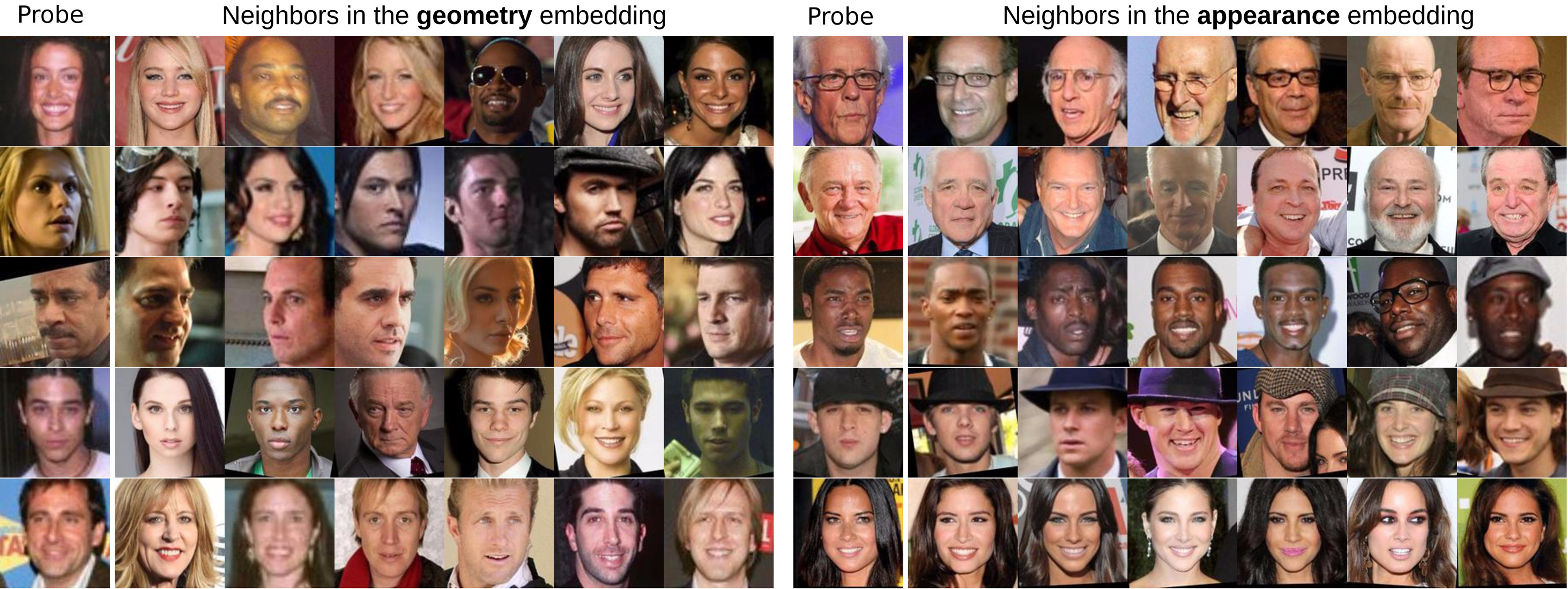}
    \caption{Visualizing the geometry (left block) and appearance (right block) embedding representations. For each probe sample, six nearest neighbors, based on cosine similarity in the embedding space, are demonstrated.}
    \label{fig:visembeddings}
\end{figure*}
\vspace{2pt}

\noindent{\bf Preprocessing:} Throughout the experiments, all faces are detected and aligned using DLib \cite{king2009dlib}. For each face, $68$ landmark locations are extracted, and the closest neighbor in the geometry space is selected using Equation \ref{eq:neighbor} over $1000$ randomly selected face images from different IDs. Neighbor faces are then transformed spatially using Equation \ref{eq:spatialtransformation}, and again aligned to compensate for the displacements caused by the spatial transformation. All face images are then resized to $112\times112$ and pixel values are scaled to $[-1, 1]$.

\subsection{Exploratory Experiments}\label{sec:lambda}
In this section, we first conduct experiments to evaluate the role of two hyper-parameters of DAG including $\lambda_a$, $\lambda_g$. Then, we perform a visualization experiment to demonstrate the effectiveness of DAG in learning rich geometry and appearance representations. 
We train a deep FR model using Softmax loss enhanced by DAG with different values of $\lambda_a$ and $\lambda_g$ in the range $[0, 2]$, and evaluate the recognition performance on the LFW \cite{huang2008labeled} dataset. Figure \ref{fig:lambdavslambda} presents the results for this experiment. The model reduces to a conventional face recognizer when $\lambda_a$ and $\lambda_g$ are zero. For large values of these parameters, the performance deteriorates which shows that the appearance and geometry loss functions dominate the identification objective. The maximum performance of the model occurs at $\lambda_a = 1.3$ and $\lambda_g=0.75$. This confirms that DAG can enhance the performance of face recognition models. Furthermore, the performance of the model is more sensitive to $\lambda_g$ compared to $\lambda_a$ which shows that matching geometry representations of $x_i$ and $\hat{x}_{i'}$ is harder than matching the appearance representations of $x_{i'}$ and $\hat{x}_{i'}$. We attribute this to the slight mismatch between the geometry of identical faces introduced because of the limited number of landmarks used to match the geometry of faces. 
On the other hand, the geometric transformation does not affect the appearance of faces. Hence, reducing the angular distance of appearance representations is more compatible with the identification loss.   

Figure \ref{fig:visembeddings} presents a visualization of the embedding space representations learned by DAG. For this purpose, nearest neighbors of several probe faces are computed in the appearance or geometry embeddings based on their cosine similarity. Inspecting neighbor faces in geometry embedding suggests that DAG robustly captures geometry information of faces, such as relative distance and sizes of parts. Also, the large appearance variations of neighbors in the geometry embedding highlights that $\g(x)$ is invariant to the appearance. On the other hand, neighbors in the appearance embedding illustrate semantic appearance characteristics such as skin color, hair color, age, and gender. Interestingly, we observe that $\a(x)$ also captures appearance properties, such as the presence of eyeglasses and a hat, which are less prevalent compared to hair color and gender.      

\subsection{Face Recognition Enhanced by DAG}\label{sec:facerecongitionenhancedbydag}
\subsubsection{Performance of Combined Loss Functions}\label{sec:combinedlosses}
In this section, we combine DAG with several well-known face recognition methods and evaluate their performance on LFW, YTF, and MegaFace Challenge 1(MF1) \cite{kemelmacher2016megaface}. We train models on the CASIA-WebFace \cite{yi2014learning} dataset with the same network architecture of modified ResNet-64 defined in Section \ref{sec:implementationdetail}. As discussed in Section \ref{sec:geoidenticalfaces}, DAG utilizes geometrically transformed faces to disentangle appearance and geometry. These transformed faces augment the training set which potentially can improve the performance of deep face recognition. Hence, to better analyze the effectiveness of disentangling we consider an additional model trained on a quasi-augmented dataset. This dataset consists of around 1M images and formed by appending 10,575 subjects from MS-Celeb-1M \cite{guo2016ms} to CASIA-WebFace. The size of the quasi-augmented dataset is equal to the presumably augmented dataset constructed by the geometric transformation of DAG.    
Table \ref{tab:combinedevaluation} demonstrates the results for these experiments. As expected, training models on quasi-augmented dataset improves the performance. However, combining face recognition models with DAG consistently outperforms baselines. This suggests that disentangling the two major characteristics of faces enhances the training process of deep models and help them learn more abstract and representative features compared to the case when solely the training set is enlarged. 
\begin{table}
    \centering
    \small
    \begin{tabular}{|c|cccc|}
    \hline
        Method & LFW & YTF & \begin{tabular}{@{}c@{}}MF1 \\ Rank1\end{tabular} & \begin{tabular}{@{}c@{}}MF1 \\ Veri.\end{tabular}  \\ \hhline{=====} 
        Softmax \cite{liu2017sphereface} &  $97.89$ & $93.1$ & $54.88$ & $66.31$\\
        Softmax+Aug. &  $98.15$ & $94.5$ & $58.90$ & $70.02$\\ 
        Softmax{\bf+DAG} &  $\boldsymbol{98.58}$ & $\boldsymbol{94.7}$ & $\boldsymbol{60.73}$ & $\boldsymbol{71.62}$\\ \hline
        SphereFace \cite{liu2017sphereface} & $99.40$ & $94.9$ & $73.19$ & $86.38$ \\
        SphereFace+Aug.  & $99.46$  & $95.2$ & $74.66$ & $88.35$ \\
        SphereFace{\bf+DAG}  & $\boldsymbol{99.55}$  & $\boldsymbol{95.6}$ & $\boldsymbol{75.28}$ & $\boldsymbol{88.90}$ \\\hline
        CosFace \cite{wang2018cosface} & $99.34$ & $95.8$ & $77.15$ & $89.76$\\
        CosFace+Aug.  & $99.48$ & $96.2$ & $78.31$ & $90.12$ \\
        CosFace{\bf+DAG}  & $\boldsymbol{99.59}$ & $\boldsymbol{97.2}$ & $\boldsymbol{79.24}$ & $\boldsymbol{91.04}$ \\\hline
    \end{tabular}
    \vspace{5pt}
    \caption{Evaluating the performance of well-known face recognition models enhanced with DAG. 
    Verification refers to true acceptance rate under FAR$=10^{-6}$. }
    \label{tab:combinedevaluation}
    \vspace{-10pt}
\end{table}
\subsubsection{Benchmark Evaluations}\label{sec:benchmarkevaluations}
For a fair benchmark comparison, we train the model on a large dataset of face images formed by combining VGGFace2 \cite{cao2018vggface2} and a private dataset. VGGFace2 contains $3.3$M images from $9.1$K identities with the average sample per identity of $362$. The final dataset encompasses $4$M images and $11.3$K identities. 

\begin{table}[]
    \centering
    \small
    \begin{tabular}{|c|c|c|c|}
    \hline
        Method & Training size & ~LFW~ & ~YTF~ \\\hhline{====}
        Deep Face \cite{taigman2014deepface} & 4M & $97.35$ & $91.4$ \\
        FaceNet \cite{schroff2015facenet} & 200M & $99.65$ & $95.1$ \\
        DeepFR \cite{parkhi2015deep} & 2.6M & $98.95$ & $97.3$ \\
        Baidu \cite{liu2015targeting} & 1.3M & $99.13$ & -\\
        SphereFace \cite{liu2017sphereface} & 0.49M & $99.42$ & $95.0$ \\
        CosFace \cite{wang2018cosface} & 5M & $99.73$ & $97.6$ \\\hline
        SphereFace{\bf+DAG} & 4M & $99.67$ & $96.2$ \\
        CoseFace{\bf+DAG} & 4M & $\boldsymbol{99.81}$ & $\boldsymbol{98.0}$
        \\\hline
    \end{tabular}
        \vspace{5pt}
    \caption{Benchmark evaluation of face verification performance (\%) on LFW and YTF.}
    \label{tab:benchmarklfw}
    \vspace{-10pt}
\end{table}
\begin{table}[]
    \centering
    \small
    \begin{tabular}{|c|c|c|c|}
    \hline
        Method & Protocol & Acc. & Veri. \\ \hhline{====}
        SIAT\_MMLAB \cite{wen2016discriminative} & Small & $65.23$ & $76.72$ \\
        DeepSense-Small & Small & $70.98$ & $82.85$ \\
        BeijingFaceAll V2 & Small & $76.66$ & $77.60$ \\
        GRCCV & Small & $77.67$ & $74.88$ \\
        FUDAN\_CS\_SDS \cite{wang2017multi} & Small & $77.98$ & $79.19$ \\
        SphereFace (1-patch) \cite{liu2017sphereface} & Small & $72.72$ & $85.56$ \\ 
        SphereFace (3-patch) \cite{liu2017sphereface} & Small & $75.76$ & $89.14$ \\ 
        CosFace (1-patch) \cite{wang2018cosface} & Small & $77.11$ & $89.88$ \\
        CosFace (3-patch) \cite{wang2018cosface} & Small & $79.54$ & $92.22$ \\\hline
        SphereFace{\bf+DAG} (1-patch) & Small & $77.32$ & $91.25$ \\ 
        SphereFace{\bf+DAG} (3-patch) & Small & $78.83$ & $92.24$ \\ 
        CosFace{\bf+DAG} (1-patch) & Small & $79.18$ & $91.46$ \\
        CosFace{\bf+DAG} (3-patch) & Small & $\boldsymbol{82.54}$ & $\boldsymbol{94.79}$ \\ \hhline{====}
        Beijing FaceAll\_Norm 1600 & Large & $64.80$ & $67.11$ \\
        Google-FaceNet v8 \cite{schroff2015facenet} & Large & $70.49$ & $86.47$ \\
        NTechLab-facenx large & Large & $73.30$ & $85.08$\\
        SIATMMLAB TencentVision & Large & $74.20$ & $87.27$ \\
        DeepSense V2 &Large & $81.29$& $95.99$ \\
        Youtu Lab &Large& $83.29$&$91.34$\\
        Vocord-deepVo V3 & Large & $\boldsymbol{91.76}$&$94.96$ \\
        SphereFace (1-patch) \cite{liu2017sphereface} & Large & $77.44$ & $91.49$ \\ 
        SphereFace (3-patch) \cite{liu2017sphereface} & Large & $80.85$ & $93.60$ \\ 
        CosFace (1-patch) \cite{wang2018cosface} & Large & $82.72$ & $96.65$ \\
        CosFace (3-patch) \cite{wang2018cosface} & Large & $84.26$ & $97.96$ \\  \hline
        SphereFace{\bf+DAG} (1-patch) & Large & $81.28$ & $93.32$ \\ 
        SphereFace{\bf+DAG} (3-patch) & Large & $85.76$ & $94.87$ \\ 
        CosFace{\bf+DAG} (1-patch) & Large & $85.62$ & $97.26$ \\
        CosFace{\bf+DAG} (3-patch) & Large & $87.02$ & $\boldsymbol{98.29}$ 
        
        \\\hline
    \end{tabular}
        \vspace{5pt}
    \caption{Performance of face identification and verification on MegaFace dataset. Verification measure (Veri.) denotes TAR at $FAR=10^{-6}$.}
    \label{tab:megafacebenchmark}
    \vspace{-10pt}    
\end{table}
\noindent{\bf LFW and YTF.}
For evaluating the model on LFW, we follow the standard protocol of unrestricted with labeled outside data \cite{huang2008labeled} and report our results on 6,000 pairs constructed using the test subset. YTF \cite{wolf2011face} consists of 3,425 videos of 1,595 unique IDs. Each video contains 181.3 frames on average which are downloaded from YouTube. Again, we follow the standard protocol of unrestricted with labeled outside data \cite{huang2008labeled} and conduct experiments on 5,000 video pairs. Table \ref{tab:benchmarklfw} presents the results for these experiments. Integrating DAG with the well-known face recognition methods consistently enhances their performance on both LFW and YTF datasets. 

\noindent{\bf MegaFace.} We further evaluate the identification and verification performance of face recognition models enhanced by DAG using the challenging and large-scale benchmark of MegaFace \cite{kemelmacher2016megaface}. MegaFace consists of a probe and a gallery subset. The gallery contains more than 1 million images from 640k individuals. The probe dataset is formed by combining FaceScrub \cite{ng2014data} and FGNet datasets. The first dataset contains 100K images from 530 unique IDs, and the second dataset contains 1,002 images from 82 IDs. Several standard test scenarios are defined to evaluate the identification, verification, and pose invariance performance of methods under two main protocols, \ie, small and large training sets. The protocol is considered small or large when the training set involves less than or greater than 0.5 million images, respectively. We also consider multi-patch models to measure the performance of an ensemble of the proposed model based on the setup described in \cite{liu2017sphereface}.    

Table \ref{tab:megafacebenchmark} summarizes the results for these evaluations. On both protocols, integrating DAG with SphereFace and CosFace enhances both the identification and verification performances. Particularly, on three out of four test setups, the integration with CosFace achieves superior performance compared to the previous approaches.   

\begin{figure}
    \centering
    \includegraphics[width=0.35\textwidth]{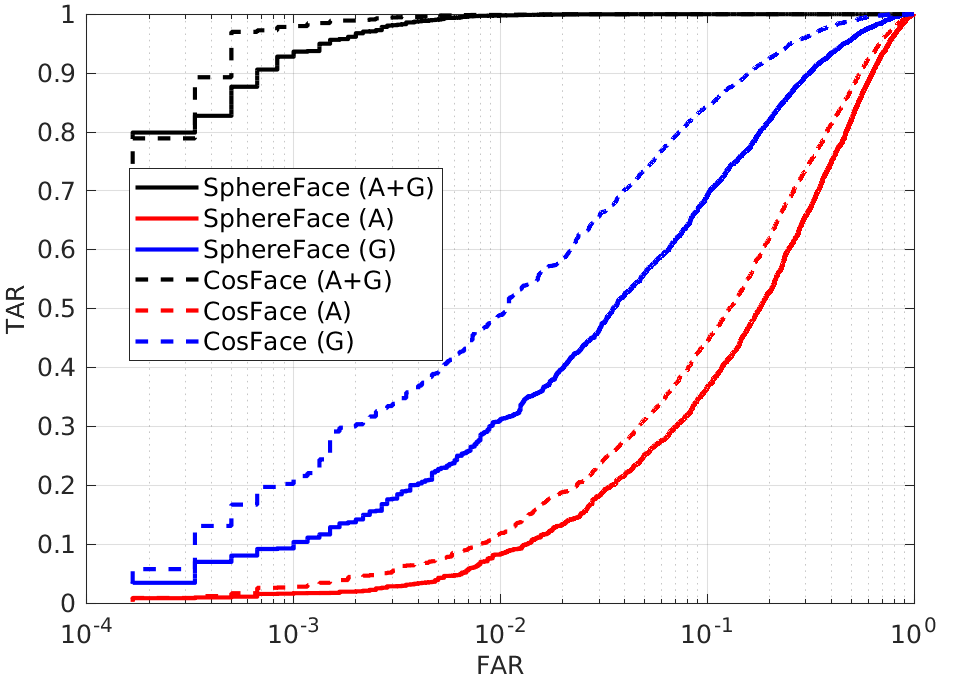}
    \caption{ROC curves for matching face images based on representations of appearance (A), geometry (G), and their combination (A+G) on LFW \cite{huang2008labeled}. }
    \label{fig:rocindividual}
\end{figure}

\begin{table}[]
    \centering
    \small
    \begin{tabular}{|c|ccc|}
    \hline
        Method & $\a(x)$ & $\g(x)$ & $f(\a(x), \g(x))$ \\\hhline{====}
        SphereFace{\bf+DAG} & $67.03$ & $81.12$ & $99.55$  \\
        CosFace{\bf+DAG} & $68.56$ & $87.45$ & $99.59$ 
         \\\hline
    \end{tabular}
    \caption{Identification performance of individual representations on LFW \cite{huang2008labeled}.}
    \label{tab:individualperformance}
    \vspace{-6pt}
\end{table}

\begin{table*}[]
    \centering
    \footnotesize
    \setlength\tabcolsep{2.5pt}
    \begin{tabular}{|c|cccccccccccccccccc|}
    \hline
    Method &\rotatebox[origin=l]{90}{Bald} & \rotatebox[origin=l]{90}{Bangs} & \rotatebox[origin=l]{90}{Big Lips} &
        \rotatebox[origin=l]{90}{Big Nose} &
        \rotatebox[origin=l]{90}{Black Hair} &
        \rotatebox[origin=l]{90}{Blond Hair} &
        \rotatebox[origin=l]{90}{Brown Hair} &
        \rotatebox[origin=l]{90}{Bushy Eyeb.} &
        \rotatebox[origin=l]{90}{Chubby}&
        \rotatebox[origin=l]{90}{Eyeglasses}&
        \rotatebox[origin=l]{90}{Male}&
        \rotatebox[origin=l]{90}{Mustache}&
        \rotatebox[origin=l]{90}{Narr. Eye} &
        \rotatebox[origin=l]{90}{No beard} &
        \rotatebox[origin=l]{90}{Oval Face} &
        \rotatebox[origin=l]{90}{Smiling} &
        \rotatebox[origin=l]{90}{Wear. Hat}&
        \rotatebox[origin=l]{90}{Young}

        \\\hhline{===================}
        FaceTracer \cite{kumar2008facetracer} &$89$&$88$&$64$&$74$&$70$&$80$&$60$&$80$&$86$&$98$&$91$&$91$&$82$&$90$&$64$&$89$&$89$&$80$ \\
    PANDA \cite{zhang2014panda} &$96$&$92$&$67$&$75$&$85$&$93$&$77$&$86$&$86$&$98$&$97$&$93$&$84$&$93$&$65$&$92$&$96$&$84$ \\
        LNets+ANet \cite{liu2015deep} &$98$&$95$&$68$&$78$&$88$&$95$&$80$&$90$&$91$&$99$&$98$&$\boldsymbol{95}$&$81$&$\boldsymbol{95}$&$66$&$92$&$99$&$87$ \\
        Model$_A$ &$82.3$&$80.1$&$59.5$&$70.3$&$64.9$&$71.4$&$59.6$&$70.5$&$74.0$&$85.4$&$82.6$&$82.6$&$82.1$&$73.7$&$60.6$&$76.7$&$78.6$&$68.3$ \\
        Model$_A${\bf+DAG} &$89.4$&$86.6$&$67.6$&$77.7$&$75.4$&$79.3$&$66.9$&$79.2$&$87.7$&$94.2$&$90.4$&$89.3$&$85.0$&$79.2$&$68.1$&$84.2$&$92.5$&$80.7$ \\
        Model$_B$ &$97.2$&$94.6$&$70.5$&$78.3$&$91.6$&$93.7$&$76.4$&$88.7$&$93.0$&$98.2$&$98.8$&$91.5$&$85.6$&$84.6$&$72.6$&$92.6$&$97.3$&$89.0$ \\
        Model$_B${\bf+DAG} &$\boldsymbol{99.1}$&$\boldsymbol{97.3}$&$\boldsymbol{77.8}$&$\boldsymbol{85.2}$&$\boldsymbol{92.5}$&$\boldsymbol{97.4}$&$\boldsymbol{84.6}$&$\boldsymbol{94.4}$&$\boldsymbol{94.2}$&$\boldsymbol{99.3}$&$\boldsymbol{99.1}$&$93.2$&$\boldsymbol{88.6}$&$91.6$&$\boldsymbol{77.3}$&$\boldsymbol{95.2}$&$\boldsymbol{99.2}$&$\boldsymbol{93.4}$ \\\hline
    \end{tabular}
    \vspace{2pt}
    \caption{Performance comparison of facial attribute prediction methods on CelebA.}
    \label{tab:attributeresults}
    \vspace{-10pt}
\end{table*}

\subsubsection{Performance of Individual Representations}
In previous sections, we demonstrated that disentangling appearance and geometry representations of faces can enhance the recognition performance. We attribute this performance boost to the highly representative and complementary features extracted by each of the geometry and appearance branches. Indeed, forcing the model to learn disentangled embedding spaces helps the early and intermediate convolutional layers to extract more representative features. Here, we examine the appearance and geometry representations individually to evaluate their role in the recognition task. To this aim, we consider $\g(x)$, $\a(x)$, and $f(\a(x), \g(x))$ for matching face images of LFW using the setup described in Section \ref{sec:benchmarkevaluations}. Figure \ref{fig:rocindividual} and Table \ref{tab:individualperformance} present the result for this experiment. The geometry representations on both methods provide more informative representations for the identification task compared to appearance representations. This was expected since the geometry of faces contain rich discriminative information and appearance of faces has less variations intrinsically. 

\subsection{Knowledge Transfer for Attribute Prediction}\label{sec:dagforattributeprediction}
Learning rich representations for faces can be beneficial for applications other than face recognition. Another important task related to the face is attribute prediction. Lack of both training data and variations in properties of faces in annotated datasets is the major factor deteriorating the performance of facial attribute prediction models \cite{kalayeh2017improving, taherkhani2018deep}. To address these problems, a major group of methods build their models upon representations learned from large-scale face recognition datasets \cite{liu2015deep}. In this section, we transfer the knowledge learned using a face recognition method integrated with DAG to evaluate its usefulness for attribute prediction. We conduct our experiments on the widely used face attribute dataset 
of CelebA \cite{liu2015deep} which contains 10,000 identities with around 200,000 samples. Eighteen major attributes are selected for comparing the results. 

We use the exact model trained in Section \ref{sec:combinedlosses} using the Softmax loss function and drop the last two fully-connected layers, \ie preserving $\a(x)$ and $\g(x)$. Afterward, we define two test models, namely Model$_A$ and Model$_B$. In Model$_A$, we freeze $\a(x)$ and $\g(x)$, and train a fully-connected layer to map the learned representations to the final prediction of each attribute. Hence, this model mimics the weakly-supervised framework in which all layers except the last linear layer are trained solely using recognition supervision. In Model$_B$, we also fine-tune $\a(x)$ and $\g(x)$ using $0.1\times$ the learning rate of the fully-connected layers. For each attribute, a dedicated fully-connected layer is used to perform binary classification using the conventional softmax loss function. Fully-connected layers are optimized using SGD with the initial learning rate of $0.01$ and the decay rate of $0.9$ every four epochs. All models are trained for $40$ epochs. We compare our results to FaceTracer \cite{kumar2008facetracer}, PANDA \cite{zhang2014panda}, and LNets+ANet \cite{liu2015deep}. Following the setup for FaceTracer and PANDA, we use the landmark information of faces to crop all faces.

Table \ref{tab:attributeresults} summarizes the results for this experiment. Transferring the knowledge learned by the face recognition models enhanced by DAG consistently improves the performance of attribute prediction. Particularly, for Model$_A$ which is trained using a weakly-supervised scheme, integration of DAG with the original face recognizer improves the performance of attribute prediction by $\boldsymbol{8.34\%}$ on average. This confirms that the DAG approach can help deep models to capture more informative representations of the face. Furthermore, fine-tuning the trained face recognizer enhanced by DAG using the attribute classification task outperforms baselines on 16 attributes out of 18. This also demonstrates that models enhanced by DAG can provide more sophisticated knowledge compared to conventional face recognition models.      
\vspace{-5pt}
\section{Conclusion}
In this paper, we propose the disentanglement of appearance and geometry representations for the face recognition task. We demonstrate that this approach boosts the deep face recognition performance by augmenting the training set and improving the knowledge learned by early and intermediate convolutional layers. Through extensive experiments, we validate the effectiveness of the proposed approach for face recognition and facial attribute prediction on challenging datasets. The individual capacity of the appearance and geometry representations are evaluated in additional experiments to analyze their semantic role in the face recognition task. Our results suggest that task-specific considerations for the training phase can further improve the performance of deep learning models.

\clearpage
{\small
\bibliographystyle{ieee}
\bibliography{egbib}
}

\end{document}